\documentclass[10pt,twocolumn,letterpaper]{article}

\usepackage{cvpr}
\usepackage{times}
\usepackage{epsfig}
\usepackage{graphicx}
\usepackage{amsmath}
\usepackage{amssymb}

\usepackage{color}
\usepackage{verbatim}
\usepackage{amsmath}
\usepackage{booktabs}
\usepackage{graphicx}
\usepackage{amssymb}
\usepackage{multirow}

\newcommand{\object}[1]{\texttt{#1}}
\newcommand{\predicate}[1]{\texttt{#1}}
\newcommand{\relationship}[3]{$<$\texttt{#1} - \texttt{#2} - \texttt{#3}$>$}

\usepackage[pagebackref=true,breaklinks=true,letterpaper=true,colorlinks,bookmarks=false]{hyperref}

\cvprfinalcopy % *** Uncomment this line for the final submission

\def\cvprPaperID{5258} % *** Enter the CVPR Paper ID here

\ifcvprfinal\fi
\begin{document}

%%%%%%%%% TITLE
\title{Learning Predicates as Functions to Enable Few-shot Scene Graph Prediction}

\author{
Apoorva Dornadula, Austin Narcomey, Ranjay Krishna, Michael Bernstein, Li Fei-Fei \\
Stanford University\\
\texttt{\{apoorvad, anarc, ranjaykrishna, msb, feifeili\}@cs.stanford.edu} \\
}

\maketitle

\begin{abstract}
Scene graph prediction --- classifying the set of objects and predicates in a visual scene --- requires substantial training data. However, most predicates only occur a handful of times making them difficult to learn. We introduce the first scene graph prediction model that supports few-shot learning of predicates. Existing scene graph generation models represent objects using pretrained object detectors or word embeddings that capture semantic object information at the cost of encoding information about which relationships they afford. So, these object representations are unable to generalize to new few-shot relationships. We introduce a framework that induces object representations that are structured according to their visual relationships. Unlike past methods, our framework embeds objects that afford similar relationships closer together. This property allows our model to perform well in the few-shot setting. For example, applying the `riding' predicate transformation to `person' modifies the representation towards objects like `skateboard' and `horse' that enable riding. We generate object representations by learning predicates trained as message passing functions within a new graph convolution framework. The object representations are used to build few-shot predicate classifiers for rare predicates with as few as $1$ labeled example. We achieve a $5$-shot performance of $22.70$ recall@$50$, a $3.7$ increase when compared to strong transfer learning baselines.
\end{abstract}

\section{Introduction}
Scene graph prediction has shown to improve multiple Computer Vision tasks including object localization~\cite{krishna2018referring}, image captioning~\cite{anderson2016spice} and visual question answering~\cite{johnson2017inferring}. This task takes as input an image, and outputs a set of relationships denoted as \relationship{subject}{predicate}{object}, such as \relationship{woman}{drinking} {coffee} and \relationship{coffee}{on}{table}. However, due to the uneven distribution of training relationship instances in the world and in training data, existing scene graph models are only performant with the most frequent relationships (predicates). Therefore, all scene graph models to date have ignored the long tail of rare relationships.

\begin{figure}
    \centering
    \includegraphics[width=\columnwidth]{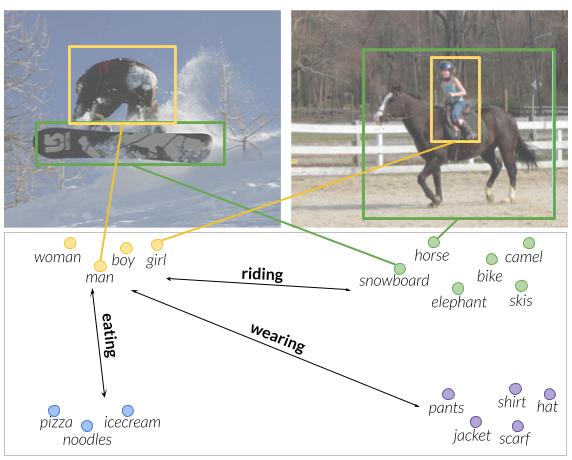}
    \caption{We introduce a new scene graph model which generates object representations that capture the visual relationships objects participate in. These representations are learned in a graph convolution network and can afford few-shot prediction of predicates, enabling models to predict rare relationships.}
    \label{fig:pull_figure}
\end{figure}

To enable a model to learn predicates with few examples, we need a mechanism to create object representations that encode the relationships afforded by the object. If we transform a subject representation by a specific predicate, the resulting object representation should be close to other objects that afford similar relationships with the subject. For example, if we transform the subject, \object{person}, with the predicate \predicate{riding}, the resulting object representation should be close to the representations of objects that can be ridden.

Existing scene graph models~\cite{ gu2019scene, lu2016visual, zellers2017neural, chen2019scene, xu2017scene} are unable to perform few-shot prediction of predicates because their object representations do not encode sufficient information about the relationships they afford. Neural Motifs~\cite{zellers2017neural}, for example, leverages linguistic priors to represent objects. Their work finds that object categories, and not their representations, are largely indicative of the relationship present between objects, thereby relying on dataset biases. However, relying on such biases doesn't transfer --- a model seeing one example of \predicate{riding} in the context of \relationship{person}{riding}{horse} doesn't learn to predict \predicate{riding} when it sees \relationship{person}{?}{snowboard}. Such few-shot transfer is only possible if \object{snowboard} and \object{horse} have similar object representations. How can we design a scene graph model that learns to encode information about visual relationships into object representations?
%Such representations can only be learned in conjunction with learning predicates and therefore, warrants a new architecture, unlike the ones already proposed by existing scene graph approaches.

We introduce a new scene graph model which generates object representations that map objects that participate in similar relationships together. Since these representations are shaped by the relationships they occur with, we need a fewer number of samples to learn new relationships. Our main insight lies in creating a new graph convolution model to learn these object representations by treating predicates as message passing functions. Each predicate function is a neural network that transforms the subject representations towards object representations for a given \relationship{subject}{predicate}{object} relationship. Similarly, the inverse predicate functions transform the object representations towards the subjects'. The forward predicate function learns to structure object representations while the inverse predicate function learns to structure the subjects'. For example, the function for \predicate{riding} learns to move to the embedding to where rideable objects like \object{horse}, \object{skateboard}, and \object{bike} can be found. Notice how traditional embeddings like GloVe~\cite{pennington2014glove} would not embed \object{horse} and \object{bike} close together. %We further divide each forward and inverse function into two components: a spatial component that transforms attention over the image space~\cite{krishna2018referring} and a semantic component that operates over the object features~\cite{zhang2017visual}.

Through our experiments on Visual Genome~\cite{krishna2017visual}, a dataset containing visual relationships and scene graphs, we show that the object representations generated by the predicate functions result in meaningful features that can be used to enable few-shot scene graph prediction. This model exceeds the GloVe~\cite{pennington2014glove} baseline by $3.7$ and the existing transfer learning approaches by $8.7$ at recall@$50$ with $5$ labelled examples. We demonstrate that our scene graph model outperforms models that also do not utilize external knowledge bases~\cite{gu2019scene}, linguistic priors~\cite{lu2016visual,zellers2017neural}, or rely on complicated pre- and post-processing heuristics~\cite{zellers2017neural,chen2019scene}. Our model performs on par with existing state-of-the-art models that do utilize this additional information. We run ablations where we remove components of our model and study how each affects performance. Since our predicates are transformation functions, we can visualize them individually, enabling the first interpretable scene graph model.

\section{Related work}

% Contextual understanding.
\paragraph{Scene graph.} Scene graphs were introduced as a formal representation for visual information~\cite{johnson2015image,krishna2017visual} in a form widely used in knowledge bases~\cite{guodong2005exploring,culotta2004dependency,zhou2007tree}. Each scene graph encodes objects as nodes connected together by pairwise relationships as edges. Scene graphs have led to many state of the art models in image captioning~\cite{anderson2016spice}, image retrieval~\cite{johnson2015image,schuster2015generating}, visual question answering~\cite{johnson2017inferring}, relationship modeling~\cite{krishna2018referring}, and image generation~\cite{johnson2018image}. Given its versatile utility, the task of scene graph prediction has resulted in a series of publications~\cite{krishna2017visual,dai2017detecting,liang2017deep,li2017vip,li2017scene,newell2017pixels,xu2017scene,zellers2017neural,yang2018graph,herzig2018mapping} that have explored reinforcement learning~\cite{liang2017deep}, structured prediction~\cite{krahenbuhl2011efficient,desai2011discriminative,tu2010auto}, utilizing object attributes~\cite{farhadi2009describing,parikh2011relative}, sequential prediction~\cite{newell2017pixels}, and graph-based~\cite{xu2017scene, li2018factorizable, yang2018graph} approaches. However, all of these approaches have classified predicates using object features, confounding the object features with predicate information that prevents their utility when used to train new few-shot predicate categories.

% Modeling differences.
\paragraph{Predicates and relationships.} The strategy of decomposing relationships into their corresponding objects and predicates has been recognized in other works~\cite{li2018factorizable,yang2018graph} but we generalize existing methods by treating predicates as functions, implemented as general neural network modules. Recent work on referring relationships showed that predicates can be learned as spatial transformations in visual attention~\cite{krishna2018referring}. We extend this idea to formulate predicates as message passing semantic and spatial functions in a graph convolution framework. This framework generalizes existing work~\cite{li2018factorizable,yang2018graph} where relationships are usually treated as latent representations instead of functions. It also generalizes papers that have restricted these functions to linear transformations~\cite{bordes2013translating,zhang2017visual}. We derive object representations to be used for few-shot predicate prediction by learning predicates as functions in a graph convolution network.

\begin{figure*}[t]
    \centering
    \includegraphics[width=\linewidth]{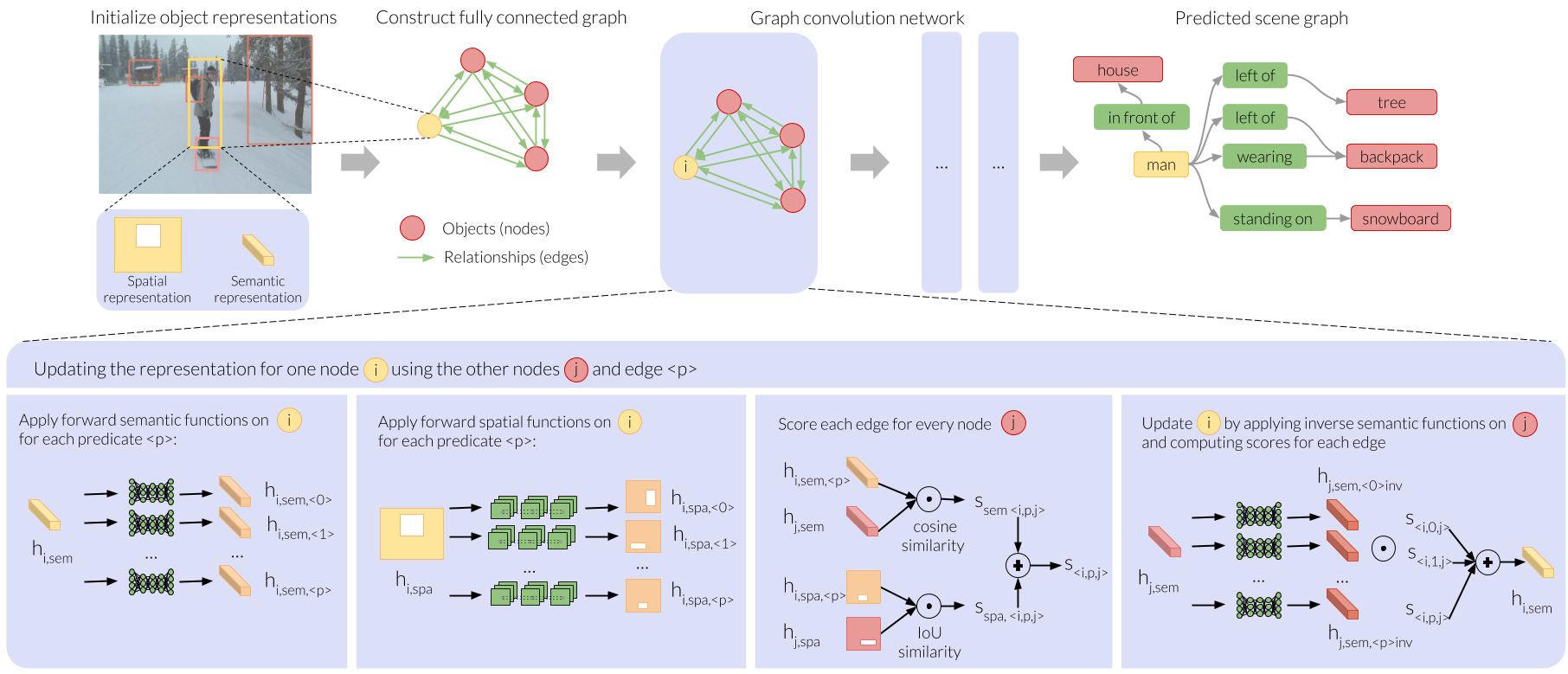}
    \caption{We introduce a scene graph prediction model that induces object representations that captures the relationships the object affords. First, we extract bounding box proposals from an input image and represent objects as semantic features and spatial attentions. Next, we construct a fully connected graph where object representations form the nodes and the predicate functions act as edges. Here we show how one node, the \object{person}'s representation is updated within one graph convolution step. Over multiple steps, the object representations produced are effective for few-shot prediction of rare predicates with as few as $1$ training example.}
    \label{fig:model}
\end{figure*}

% Graph Convolution Networks
\paragraph{Graph convolutions.} Modeling graphical data has historically been challenging, especially when dealing with large amounts of data~\cite{weston2012deep,belkin2006manifold,zhou2004learning}. Traditional methods have relied on Laplacian regularization through label propagation~\cite{zhou2004learning}, manifold regularization~\cite{belkin2006manifold}, or learning embeddings~\cite{weston2012deep}. Recently, operators on local neighborhoods of nodes have become popular with their ability to scale to larger amounts of data and parallelizable computation~\cite{grover2016node2vec,perozzi2014deepwalk}. Inspired by these Laplacian-based, local operations, graph convolutions~\cite{kipf2016semi} have become the de facto choice when working with graphical data~\cite{kipf2016semi,scarselli2009graph,li2015gated,henaff2015deep,duvenaud2015convolutional,niepert2016learning}. Graph convolutions have recently been combined with RCNN~\cite{girshick2015fast} to perform scene graph detection~\cite{yang2018graph,johnson2018image}.
Unlike most graph convolution methods, which assume a known graph structure, our framework doesn't make any prior assumptions to limit the types of relationships between any two object nodes, i.e.~we don't use relationship proposals to limit the possible edges. Instead, we learn to score the predicate functions between the nodes, strengthening the correct relationships and weakening the incorrect ones over multiple iterations. This manner of learning predicate functions allows us to embed relationship context in the object representations learned by the network.

% K-Shot Literature
\paragraph{Few-shot prediction.} While graph-based learning typically requires large amounts of training data, we extend work in few-shot prediction, to show how the object representations learned using predicate functions can be further used to transfer to rare predicates. The few-shot literature is broadly divided into two main frameworks. The first strategy learns a classifier for a set of frequent categories and then uses them to learn the few-shot categories~\cite{fe2003bayesian,fei2006one,lake2011one,snell2017prototypical,mehrotra2017generative, chen2019scene}. The second strategy learns invariances or decompositions that enable few-shot classification~\cite{koch2015siamese,vinyals2016matching,triantafillou2017few,garcia2017few}. Our framework more closely resembles the first framework.

% Modular networks.
\paragraph{Modular neural networks.} Utilizing the modularity of neural networks have shown to be useful in numerous machine learning applications~\cite{andreas2016neural,kumar2016ask,xiong2016dynamic,andreas2016learning,johnson2017inferring}. Typically, their utility has focused on the ability to train individual components and then jointly fine-tune them. Our paper focuses on a complementary ability of such networks: our functions are trained together and then used to learn additional predicates without retraining the entire model.

\section{Graph convolution framework with predicate functions}
In this section, we describe our graph convolution framework (Figure~\ref{fig:model}) and the predicate functions. This framework is responsible for creating the object representations using frequent predicates in a graph convolution framework. The representations will used in the next section to enable few-shot prediction of rare predicates. 
%In the next section, we use the representations generated by these functions to train few-shot predicate classifiers.

\subsection{Problem formulation}
Our goal is to learn an effective object representations using frequent predicates. To ensure that the representation projects objects that participate in similar relationships together, we design the predicates as functions that transform object embeddings within a graph convolution network.These functions are learned during for the task of scene graph generation. Formally, the input to our model is an image $I$ from which we extract a set of bounding box proposals $B = \{b_1, b_2, \ldots\, b_n\}$ using a region proposal network~\cite{ren2015faster}. From these bounding boxes, we extract initial object features $H^0 = \{h_1^0, h_2^0, \ldots\, h_n^0\}$. These boxes and features are sent to our graph convolution framework. The final output of our model is a scene graph denoted as $G = \{\mathcal{V}, \mathcal{E}, \mathcal{P}\}$ with nodes (objects) $v_i \in \mathcal{V}$, and labeled edges (relationships) $e_{ijp} = <v_i, p, v_j> \in \mathcal{E}$, where $p\in\mathcal{P}$ is one of $\lvert \mathcal{P} \rvert$ predicate categories.

\subsection{Traditional graph convolutional network}
Our model is primarily motivated as an extension to graph convolutional networks that operate on local graph neighborhoods~\cite{duvenaud2015convolutional,schlichtkrull2017modeling,kipf2016semi}. These methods can be understood as simple message passing frameworks~\cite{gilmer2017neural}:
\begin{align}
m_i^{t+1} = \sum_{j \in N(i)} M(h_i^t, h_j^t, e_{ij}), \;\;\;\;\;
h_i^{t+1} = U(h_i^t, m_i^{t+1}) 
\end{align}
where $h_i^t$ is a hidden representation of node $v_i$ in the $t^{th}$ iteration, $M$ and $U$ are respectively aggregation and vertex update functions that accumulate information from the other nodes. $N(i)$ is the set of neighbors of $i$ in the graph.

\subsection{Our graph convolutional network}
Similar to previous work~\cite{schlichtkrull2017modeling} which used multiple edge categories, we expand the above formulation to support multiple edge types, i.e.~given two nodes $v_i$ and $v_j$, an edge exists from $v_i$ to $v_j$ for all $\lvert \mathcal{P} \rvert$ predicate categories. Unlike previous work where edges are an input~\cite{schlichtkrull2017modeling}, we initialize a fully connected graph, i.e.~all objects are connected to all other objects by all predicate edges. If after the graph messages are passed, predicate $p$ is scored above a hyperparameter threshold, then that relationship $<v_i, p, v_j>$ is part of the generated scene graph. The updated equations are then,
\begin{align}
m_i^{t+1} & = \sum_{p \in \mathcal{P}} \sum_{j \neq i} M_p(h_i^t, h_j^t, e_{ijp}),\\
h_i^{t+1} & = U(h_i^t, m_i^{t+1}) = \sigma (W_0 h_i^t + m_i^{t+1})
\label{eq:update}
\end{align}
where $M_p(\cdot)$ are learned message functions between two nodes for the predicate $p$, which we will detail later in this section. Note that this formula is a generalized version of the exact representation used in the previous work~\cite{schlichtkrull2017modeling}:
\begin{align}
    M_p(h_i^t, h_j^t, e_{ijp}) = \begin{cases}
        \frac{1}{c_{i,p}}W_p h_j^t \;\;\;\textrm{if}\;\; (v_i, p, v_j) \in \mathcal{E}\\
        0 \;\;\; \textrm{otherwise},
    \end{cases}
\end{align}
and $\sigma$ is the sigmoid activation. Here, $c_{i, p}$ is a normalizing constant for the edge $(i, j)$ as defined in previous work~\cite{schlichtkrull2017modeling}.

\subsection{Node hidden representations}
With the overall update step for each node defined, we now explain the hidden object representation $h_i^t$. Traditionally, object nodes in graph models are defined as being a $D$-dimensional representation of the node $h_i \in \mathcal{R}^{D}$~\cite{duvenaud2015convolutional,schlichtkrull2017modeling,kipf2016semi}. However, in our case, we want these hidden representations to encode both the semantic information for each object proposal as well as its spatial location in the image. Although the spatial representation alone is not enough to effectively predict predicates, it allows us to learn the alignment between objects for each relationship. These two components will be separately utilized by the predicate functions. Instead of asking our model to learn to represent both of these pieces of information, we build invariances into our representation such that it knows to encode them both explicitly. Specifically, we define each hidden representation as a tuple of two entries: $h_i^t = (h_{i,sem}^t, h_{i,spa}^t)$ --- a semantic object feature $h_{i,sem}^t \in \mathcal{R}^D$ and a spatial attention map over the image $h_{i,spa} \in \mathcal{R}^{L\times L}$. In practice, we extract $h_{i,sem}^0$ from the penultimate layer in ResNet-50~\cite{He2015} and set $h_{i,spa}$ as a $L\times L$ mask with $1$ for the pixels within the object proposal and $0$ outside.

With the semantic and spatial separation, we rewrite equation~\ref{eq:update} as:
% and $U_p^l(\cdot)$. For simplicity, we define $m_{i,sem}^{l+1}$, $m_{i,spa}^{l+1}$, $M_{sem}^l(\cdot)$ and $M_{spa}^l(\cdot)$ such that:
\begin{align}
m_i^{t+1} &= (m_{i,sem}^{t+1}, m_{i,spa}),\nonumber\\
m_{i,sem}^{t+1} &= \sum_{p \in \mathcal{P}} \sum_{j \neq i} M_{sem}(h_{i,sem}^t, h_{j,sem}^t, e_{ijp})
\end{align}
Note that $m_{i, spa}$ does not get updated because we fix the object masks for each object. 

\begin{figure*}[t]
    \centering
    \includegraphics[width=\linewidth]{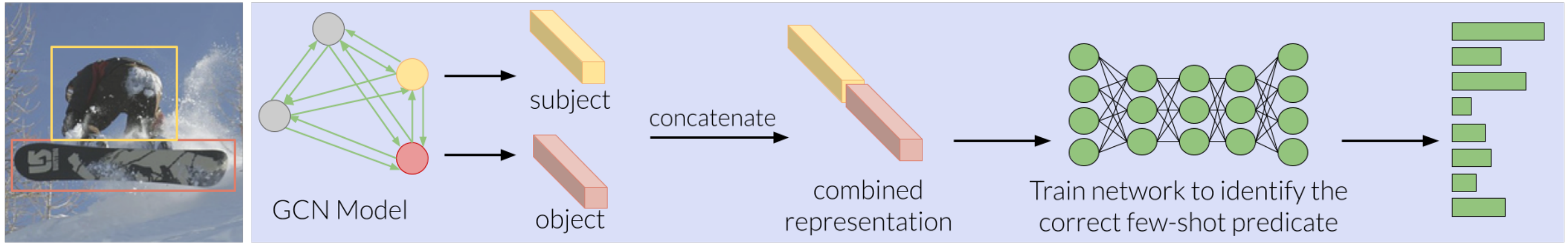}
    \caption{Overview of our few-shot training framework. We use the object representations learned from the frequent predicates in our graph convolution framework. These representations are used to train few-shot rare predicate classifiers.}
    \label{fig:k_shot}
\end{figure*}

\subsection{Predicate functions} 
To define $M_{sem}(\cdot)$, we introduce the semantic ($f_{sem,p}$) and spatial ($f_{spa,p}$) predicate functions for each predicate $p$. Semantic functions are multi-layer perceptrons (MLP) while spatial functions are convolution layers, each with $6$ layers and ReLU activations. Previous work on multi-graph convolutions~\cite{schlichtkrull2017modeling} assumed that they had a priori information about the structure of the graph, i.e.~which edges exist between any two nodes. In our case, we are attempting to perform both node classification as well as edge prediction simultaneously. Without knowing which edges actually exist, we would be adding a lot of noise if we allowed every predicate to equally influence another node. To circumvent this issue, we first calculate a score for each predicate $p$:
\begin{align}
s_{p}(h_i^t, h_j^t) &= \alpha s_{p,sem}(h_{i,sem}^t, h_{j,sem}^t) +\nonumber\\
    (1-\alpha) &s_{p,spa}(h_{i,spa}, h_{j,spa}),\\
s_{p,sem}(h_{i,sem}^t, h_{j,sem}^t) &= \mathrm{cos} \big[  f_{sem,p}(h_{i,sem}^t), h_{j,sem}^t \big],\\
s_{p,spa}(h_{i,spa}, h_{j,spa}) &= \mathrm{IoU} \big[ f_{spa, p}(h_{i,spa}), h_{j,spa} \big],
\end{align}
where $\alpha\in[0,1]$ is a hyperparameter, $\mathrm{cos}(\cdot)$ is the cosine distance function, and $\mathrm{IoU}(\cdot)$ is the differentiable intersection over union function that measures the similarity between two soft heatmaps. This gives us a score for how likely the node $v_i$ believes that the edge $<v_i, p, v_j>$ exists. Inspired by recent work~\cite{krishna2018referring}, we design $f_{spa, p}(\cdot)$ to shift the spatial attention from $h_{i,spa}$ to where it thinks node $v_j$ should be. It encodes the spatial properties of the predicate we are learning and ignores the object features. To complement the spatial predicate function, we use $f_{sem, p}(\cdot)$ to transform $h_{i, sem}^t$. This shifted representation is what the model expects to be similar to $h_{j, sem}^t$. By using both the spatial and semantic score in our update of $h_i$, the two representations interact with one another. So, even though these components are separate, they create a cohesive score for each predicate. This score is used to weight how much node $v_j$ will influence node $v_i$ through a predicate $p$ in the update in equation~\ref{eq:update}. We can now define:
\begin{align}
M_{sem}(h_{i,sem}^t, h_{j,sem}^t, e_{ijp}) = s_{p}^l(h_i^t, h_j^t) f^{-1}_{sem,p^{-1}}(h_{j,sem}^t)
\end{align}
$f_{p^{-1}}(\cdot)$ represents the backward predicate function from \object{object} back to the \object{subject}. For example, given the relationship \relationship{person}{riding}{snowboard}, our model not only learns how to transform \object{person} using the function \predicate{riding}, but also how to transform \object{snowboard} to \object{person} by using the inverse predicate $\predicate{riding}^{-1}$. 
%We update $M_{sem}$ by multiplying the score of the forward predicate by the object transformed by the inverse predicate to learn both the forward and inverse predicate functions. 
Learning both the forward and backward functions per predicate allows us to pass messages in both directions even though our predicates are directed edges.
%strengthens the signal of a particular predicate and improves relationship prediction. It also allows us two separate modes to interpret how well we model predicates.

\subsection{Hidden representation update}
$U_{sem}(\cdot)$ accumulates the messages passed by the semantic functions to update the semantic object representation: 
\begin{align}
U_{sem}(h_{i,sem}^t, m_{i,sem}^{t+1}) &= W_0 h_{i,sem}^t + \frac{1}{\lvert \mathcal{P} \rvert (\lvert \mathcal{V} \rvert -1)} m_{i,sem}^{t+1} \\
h_i^{t+1} &= (U_{sem}(h_{i,sem}^t, m_{i,sem}^{t+1}), h_{i,spa})
\end{align}
where $W_0$ is learned weight. The spatial representation does not get updated because the spatial location of an object does not move.

\subsection{Scene graph output}
We predict the node categories using $v_i = g(h_i)$, where $g$ is an MLP that generates a probability distribution over all the possible object categories. Each possible relationship $e_{ijp}$ is output as a relationship only if $s_{p}^T(h_i^T, h_j^T) * s^{-T}_{p^{-1}}(h_j^T, h_i^T) > \tau$, where $T$ is the total number of iterations in the model and $\tau$ is a threshold hyperparameter.

\begin{figure*}
    \centering
    \includegraphics[width=\linewidth]{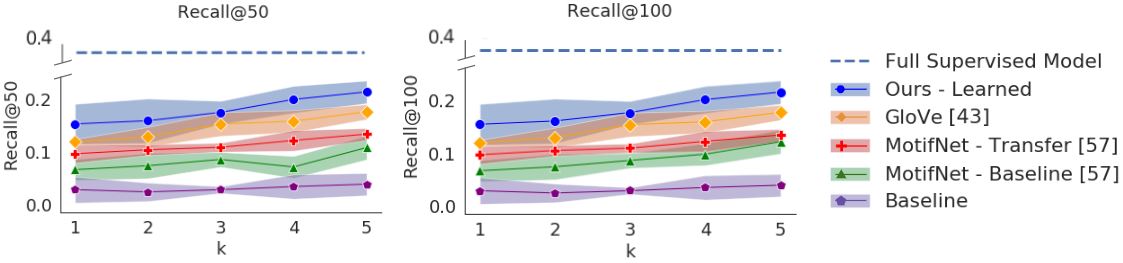}
    \caption{We show Recall@$1$ and Recall@$50$ results on $k$-shot predicates. We outperform strong baselines like transfer learning on MotifNet~\cite{zellers2017neural}, which also relies on linguistic priors.}
    \label{fig:kshot_recall}
\end{figure*}

\begin{figure*}
    \centering
    \includegraphics[width=0.9\linewidth]{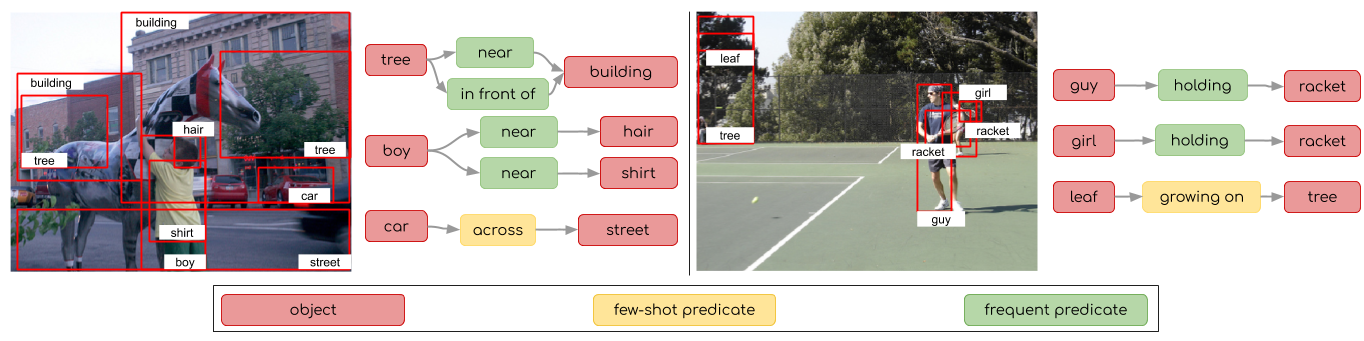}
    \caption{We show example scene graphs predicted with frequent as well as rare relationships, a feat previous models have not tackled.}
    \label{fig:kshot_outputs}
\end{figure*}

\section{Few-shot predicate framework}
%%% Put this here to push tables up higher
% Comparison with other works

% Note: our SG GEN, SG CLS, PRED CLS corresponds to SG GEN, PHR CLS, PRED CLS
\begin{table*}[t]
\centering
\caption{Our main is to enable few-shot predicate classification but also also report our performance on the general task of scene graph prediction. We perform on par with all existing state-of-the-art scene graph approaches and even outperform other methods that only utilize Visual Genome's data as supervision. We also report ablations by separating the contribution of the semantic and the spatial components.}
\resizebox{\linewidth}{!}{%
\begin{tabular}{@{\extracolsep{4pt}}l l c c c c c c}%{p{.3cm} p{3.5cm} p{1.8cm}p{1.8cm} p{1.8cm}p{1.8cm} p{1.8cm}p{1.8cm}}
  & & \multicolumn{2}{c}{\textbf{SG GEN}} & \multicolumn{2}{c}{\textbf{SG CLS}} & \multicolumn{2}{c}{\textbf{PRED CLS}}  \\
  \cmidrule(rl){3-4} \cmidrule(rl){5-6} \cmidrule(rl){7-8}
  & \textbf{Metric} & recall@50 & recall@100 & recall@50 & recall@100 & recall@50 & recall@100\\
  \midrule
  \parbox[t]{2mm}{\multirow{5}{*}{\rotatebox[origin=c]{90}{vision only}}} 
  & IMP~\cite{xu2017scene} & 06.40 & 08.00 & 20.60 & 22.40 & 40.80 & 45.20\\
  & MSDN~\cite{li2017scene} & 07.00 & 09.10 & 27.60 & 29.90 & 53.20 & 57.90\\
  & MotifNet-freq~\cite{zellers2017neural} & 06.90 & 09.10 & 23.80 & 27.20 & 41.80 & 48.80\\
  & Graph R-CNN~\cite{yang2018graph} & 11.40 & \textbf{13.70} & \textbf{29.60} & \textbf{31.60} & 54.20 & \textbf{59.10}\\
  & \textbf{Our full model} & \textbf{13.18} & 13.45 & 23.71 & 24.66 & \textbf{56.65} & 57.21\\
  \midrule
  \parbox[t]{2mm}{\multirow{4}{*}{\rotatebox[origin=c]{90}{external}}} 
  & Factorizable Net~\cite{li2018factorizable} & 13.06 & 16.47 & - & - & - & -\\
  & KB-GAN~\cite{gu2019scene} & 13.65 & 17.57 & - & - & - & -\\
  & MotifNet~\cite{zellers2017neural} & \textbf{27.20} & \textbf{30.30} & 35.80 & 36.50 & \textbf{65.20} & \textbf{67.10} \\
  & PI-SG~\cite{herzig2018mapping} & - & - & \textbf{36.50} & \textbf{38.80} & 65.10 & 66.90\\
  \midrule
  \parbox[t]{2mm}{\multirow{3}{*}{\rotatebox[origin=c]{90}{ablations}}}
  & Our spatial only & 02.05 & 02.32 & 03.92 & 04.54 & 04.19 & 04.50\\
  & Our semantic only & 12.92 & 12.39 & 23.35 & 24.00 & 56.02 & 56.67\\
  & Our full model & \textbf{13.18} & \textbf{13.45} & \textbf{23.71} & \textbf{24.66} & \textbf{56.65} & \textbf{57.21}
\end{tabular}}
\label{tab:model_gc_results_comparison}
\end{table*}

\begin{figure*}
    \centering
    \includegraphics[width=\linewidth]{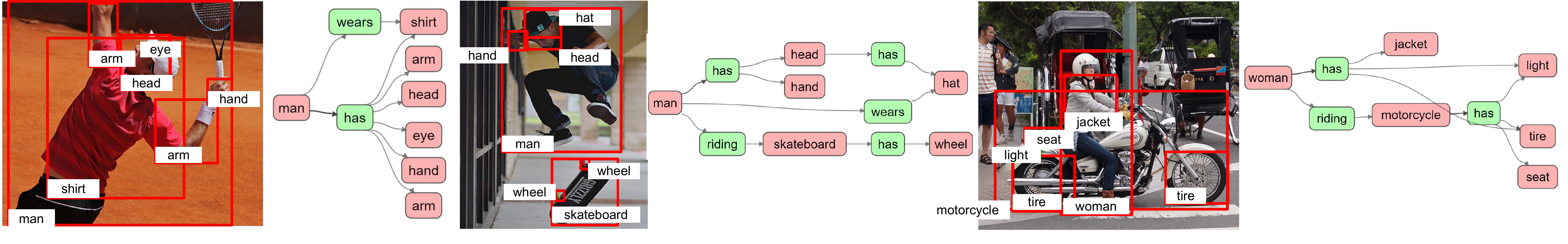}
    \caption{Example scene graphs predicted by our graph convolution fully-trained model.}
    \label{fig:example_sg}
\end{figure*}

\begin{figure*}
    \centering
    \includegraphics[width=\linewidth]{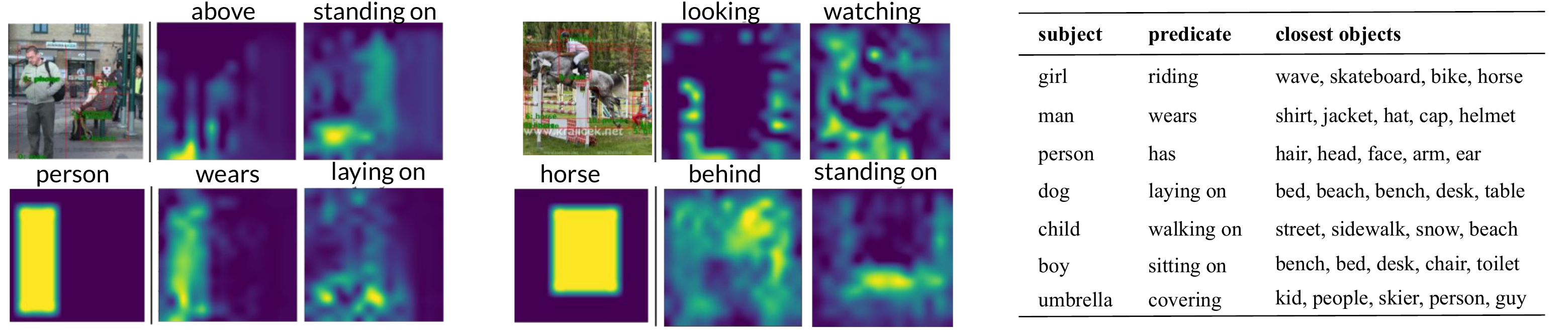}
    \caption{(left, middle) Spatial transformations learned by our model applied to object masks in images. (right) Semantic transformations applied to the average object category embedding; we show the nearest neighboring object categories to the transformed subject.}
    \label{fig:gcn_sem_spa}
\end{figure*}

Datasets used to model predicates, such as Visual Genome, have a long-tail of relationships with very few samples (few-shot), and are therefore mostly ignored by by past work, which only operate over a handful of frequent predicates. We utilize the object representations trained within the graph convolution network to train few-shot predicates. Recall that the the semantic ($f_{sem,p}$) and spatial ($f_{spa,p}$) predicate functions were trained using the frequent predicates $p \in \mathcal{P}$. We design few-shot predicate classifiers to be MLPs with $2$ layers with ReLU activations between layers. We assume that rare predicates are $p^\prime \in \mathcal{P^\prime}$ and only have $k$ examples each.

The intuition behind our $k$-shot training scheme lies in the modularity of predicates and their shared semantic and spatial components. By decomposing the predicate representations from the object in the graph convolutions, we create an representation space that supports predicate transformations. We will show in our experiments that our representation places semantically similar objects that participate in similar relationships together. Now, when training with few examples of rare predicates, such as \predicate{driving}, we can rely on the semantic embeddings for objects that were clustered by \predicate{riding}.
% and the spatial attention transformations clusters by frequent predicates like \predicate{inside of}. 

We pass all $k$ labelled examples of a predicate pair of objects $<v_i, p^\prime, v_j>$ through the learned predicate functions and extract the hidden representations $(h_{i,sem}, h_{i,spa})$ and $(h_{j,sem}, h_{j,spa})$ from the final graph convolution layer. We concatenate these transformations along the channel dimension and feed them as an input to the few-shot classifiers. We train the $k$-shot classifiers by minimizing the cross-entropy loss against the $k$ labelled examples amongst $\lvert \mathcal{P^\prime} \rvert$ rare categories.

\section{Experiments}
We begin our evaluation by first describing the dataset, evaluation metrics, and baselines. Our first experiment tests the utility of our approach on our main objective of enabling few-shot scene graph prediction. Our model outperforms competitive baselines in this task, validating our hypothesis that building object representations that capture the relationships afforded will enable better few-shot transfer. Our second experiment studies our graph convolution framework and compares our scene graph prediction performance against existing state-of-the-art methods. Recall that our aim was not to optimize scene graph prediction performance. However, our model outperforms existing models using visual features for recall@50, but as expected, do not perform well against models that utilize lisguistic priors or dataset biases. We also show that by encoding relationship information into object representations also hurt object classification. Our third experiment showcases how our model is interpretable by visualizing the predicate transformations.

\noindent\textbf{Dataset:} We use the Visual Genome~\cite{krishna2017visual} dataset for training, validation and testing. To benchmark against existing scene graph approaches, we use the commonly used subset of $150$ object and $50$ predicate categories~\cite{xu2017scene,zellers2017neural,yang2018graph}. We use publicly available pre-processed splits of train and test data, and sample a validation set from the training set~\cite{zellers2017neural}. The training, validation, and test sets contain $36,662$ and $2,794$ and $15,983$ images, respectively.

\noindent\textbf{Evaluation metrics:} For scene graph prediction, we use three evaluation tasks, all of which are evaluated at recall@$50$ and recall@$100$. (1) \texttt{PredCls} predicts predicate categories, given ground truth bounding boxes and object classes, (2) \texttt{SGCls} predicts predicate and object categories given ground truth bounding boxes, and (3) \texttt{SGGen} detects object locations, categories and predicate categories.

\noindent Metrics based on recall require ranking predictions. For \texttt{PredCls} this means a simple ranking of predicted predicates by score. For \texttt{SGCls} we ranking subject-predicate-object tuples by a product of subject, object, and predicate scores. For \texttt{SGGen} this means a similar product as SGCls, but tuples without correct subject or object localizations are not counted as correct. We refer readers to previous work that defined these metrics for further reading~\cite{lu2016visual}.

For few-shot prediction, we report recall@$1$ and recall@$50$ on the task of \texttt{PredCls}. We vary the number of labeled examples available for training few-shot predicate classifiers from $k\in [1, 2, 3, 4, 5]$. We report recall@$50$ and recall@$100$ in the test set.

\subsection{Few-shot prediction}
\noindent\textbf{Setup:}
Our first experiment studies how well we perform few-shot scene graph prediction with limited examples per predicate. Our approach requires a set of frequently occurring predicates and a set of rare predicates with only $k$ examples. We split the usual $50$ predicates typically used in Visual Genome, and place the $25$ most predicates with the most training examples into the first set and place the remaining $25$ predicates into the second set. We train the predicate functions and the graph convolution framework using the predicates in the first set. Next, we use them to train $k$-shot classifiers for the rare predicates in the second set by using the representations generated by the pretrained predicate functions. We iterate over $k \in [1, 2, 3, 4, 5]$.

\noindent\textbf{Baselines:}
In this setting, we compare against simple baselines such as using GloVe~\cite{pennington2014glove} word embeddings to represent our objects. For a rigorous comparison, we also choose to compare our method against \texttt{MotifNet}~\cite{zellers2017neural}, which outperforms all existing scene graph approaches and uses linguistic priors from word embeddings and heuristic post-processing to generate high-quality scene graphs. We report MotifNet: \texttt{MotifNet-Baseline}, which is initialized with random weights and trained only using $k$ labelled examples and \texttt{MotifNet-Transfer}, which is first trained on the frequent predicates and then finetuned on the $k$ few-shot predicates. We also compare against \texttt{GCN-Baseline}, which trains our graph convolution framework on the $k$ few-shot predicates and \texttt{Full Supervised Model}, which reports the upper bound performance when trained with all of Visual Genome.

\noindent\textbf{Results:}
in Figure~\ref{fig:kshot_recall} show that our method outperforms all baseline comparisons for all values of $k$. Our improvements increase as $k$ increases to $k=5$, where we outperform the GloVe~\cite{pennington2014glove} baseline by $3.7$ and the transfer learning baseline by $8.7$ recall@$50$. The Neural Motif model outperforms our model for values of $k \ge 10$. This is expected because Neural Motif uses the biases in the dataset between objects and predicates which become stronger as more data is used to train the model. We display scene graphs generated by our complete model with both frequent as well as rare predicates in Figure~\ref{fig:kshot_outputs}. Our performance in the few-shot learning setting validates our hypothesis that building object representations that also embed relationship information lead to better transfer to rare predicates.

\subsection{Scene graph prediction}
\noindent\textbf{Baselines:} classify existing scene graph generation methods into two categories. The first category includes other scene graph approaches that, like our approach, only utilizes Visual Genome's data as supervision. This includes Iterative Message Passing (\texttt{IMP})~\cite{xu2017scene}, Multi-level scene Description Network (\texttt{MSDN})~\cite{li2017scene}, \texttt{ViP-CNN}~\cite{li2017vip}, \texttt{MotifNet-freq}~\cite{zellers2017neural}. The second category includes models such as \texttt{Factorizable Net}~\cite{li2018factorizable}, \texttt{KB-GAN}~\cite{gu2019scene} and \texttt{MotifNet}~\cite{zellers2017neural}, which use linguistic priors or external information from knowledge bases while \texttt{MotifNet} also deploys a custom trained object detector, class-conditioned non-maximum suppression, and heuristically removes all object pairs that do not overlap. While not comparable, we report their numbers for clarity.

\noindent\textbf{Results:}
We report scene graph prediction numbers on Visual Genome~\cite{krishna2017visual} in Table~\ref{tab:model_gc_results_comparison}. This experiment is meant to serve as a benchmark against existing scene graph approaches. We outperform existing models that only use Visual Genome supervision for \texttt{SGGen} and \texttt{PredCls} by $1.78$ and $1.82$ recall@$50$, respectfully. But we fall short on recall@$100$. As we move from recall@$50$ to recall@$100$, models are evaluated on their top $100$ predictions instead of their top $50$. Unlike other models that perform a multi-class classification of predicates for every object pair, we assign binary scores to each possible predicate between an object pair individually. Therefore, we can report multiple relationships between any two objects, inflating the number of predictions our model makes, as compared to others. While this design decision allows us to separate learning predicates transformations and object representations, it penalizes our model, thereby, reducing our recall@$100$ scores. We also notice that since our model doesn't utilize the object categories to make relationship predictions, it performs worse for the task of \texttt{SGCls}, a task variant that presents models with ground truth object locations. Recall that knowing the object categories allows other models to rely on the dataset bias to guess the correct predicate~\cite{zellers2017neural}.

We report ablations of our model trained using only the semantic or spatial functions. We observe that different ablations of the model perform better on certain types of predicates. The spatial model performs well on predicates that have a clear spatial or location-based aspects, such as \predicate{above} and \predicate{under}. The semantic model performs better on non-spatial predicates such as \predicate{has} and \predicate{holding}. Our full model outperforms the semantic-only and spatial-only models as predicates can utilize both components. We visualize scene graphs generated by our network in Figure~\ref{fig:gen_sg}. Additional ablations can be found in the appendix.

\noindent\textbf{Limitations.}
One of the limitations of our framework is its computational cost. Our graph convolution framework learns object representations by learning each predicate as a function, resulting in $~1$M parameters to be learnt per predicate. Given such a large parameter space to learn, our model takes $20$ epochs to train on the Visual Genome dataset, where each forward propagation takes up to $500$ms using Titan X GPUs. Although the computational cost of our model is high, it enables us to produce effective object representations which leads to state-of-the-art performance on the few-shot scene graph generation task.

\subsection{Interpretable predicate transformations}
Our final experiment showcases another utility of treating predicates as functions. Once trained, these functions can be individually visualized and qualitatively evaluated. Figure~\ref{fig:gcn_sem_spa} (left and middle) shows examples of transforming spatial attention from four instances of \object{person}, \object{horse}, \object{boy}, and \object{banana} in four images. We see that \predicate{above} and \predicate{standing on} moves attention below the person \predicate{looking} moves attention left towards the direction the \object{horse} is looking. Figure~\ref{fig:gcn_sem_spa}(right) shows semantic transformations applied to the embedding representation space of objects. We see that \predicate{riding} transforms the embedding to a space that contains objects like \object{wave} and \object{horse}. Unlike linguistic word embeddings, which are trained to place words found in similar contexts together, our embedding space represents the types of visual relationships that objects participate. For example, \object{snow}, \object{sidewalk}, \object{beach} are mapped together since they all afford \predicate{walking on}.
% \predicate{wearing} highlights the center of the \object{boy}.

\section{Conclusion}
We introduced the first few-shot predicate prediction model, which uses effective object representations that capture how objects afford specific visual relationships. We treat frequent predicates as neural network transformations between object representations within a graph convolution network. The functions help learn a representation that embeds objects that afford similar relationships close together. The few-shot classifiers trained using the representations outperform existing methods for few-shot predicate prediction, a valuable task since most predicates occur infrequently. Our graph convolution network, performs on par with existing scene graph prediction state-of-the-art models. Finally, the predicate functions result in interpretable visualizations, allowing us to visualize the spatial and semantic transformations learned for each predicate.

\noindent\textbf{Acknowledgements}  We thank Iro Armeni, Suraj Nair, Vincent Chen, and Eric Li for their helpful comments. This work was partially funded by the Brown Institute of Media Innovation and by Toyota Research Institute (“TRI”) but this article solely reflects the opinions and conclusions of its authors and not TRI or any other Toyota entity.

{\small
\bibliographystyle{ieee_fullname}
\bibliography{egbib}
}

\section{Appendix}
We include additional scene graph outputs by our graph convolution model, add more ablations for the model, include visualizations for the spatial and the semantic transformations, and finally, plot a visualization of the object feature space.

\subsection{More scene graph model outputs}
\begin{figure*}[t!]
    \centering
    \includegraphics[width=0.9\linewidth]{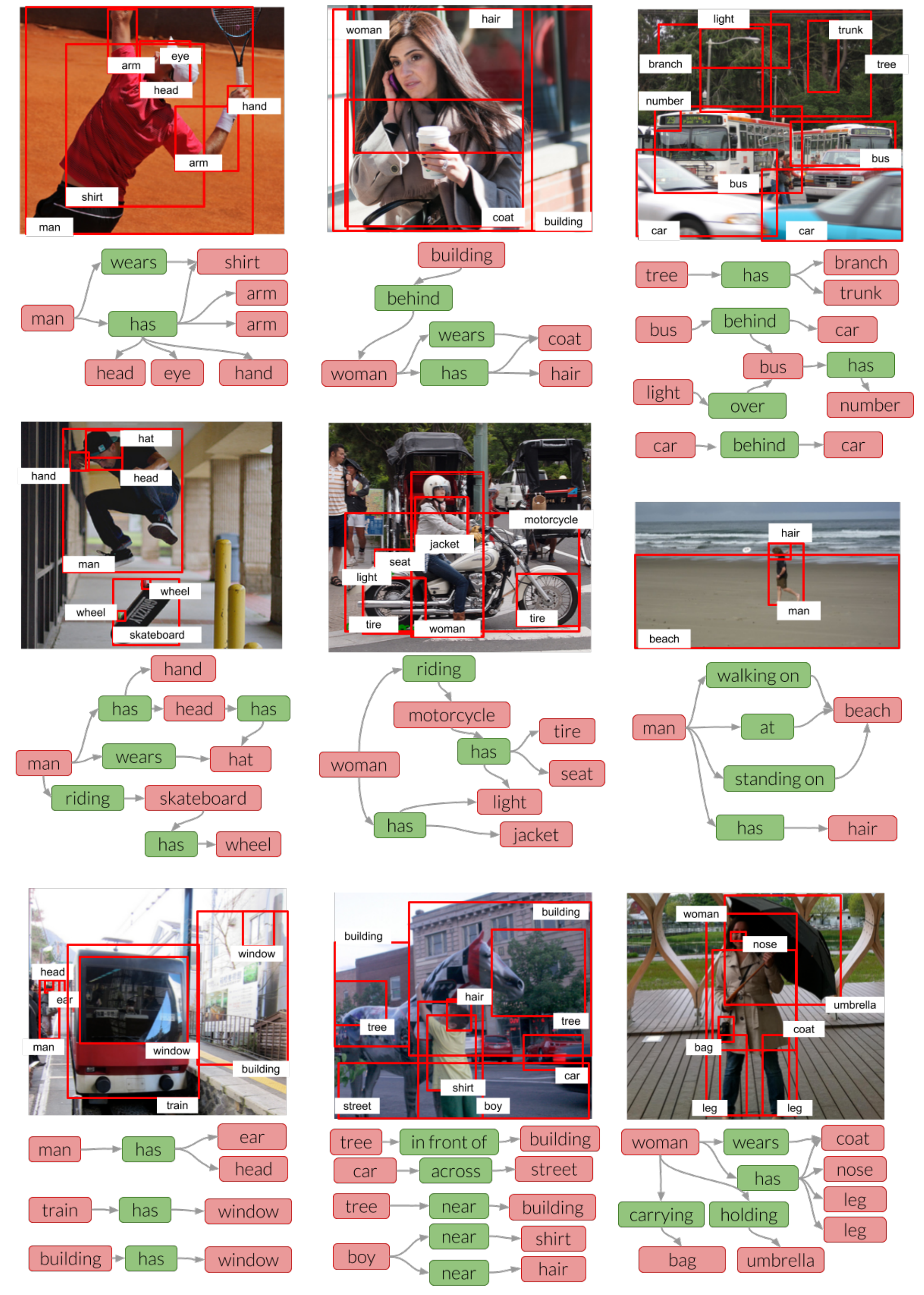}
    \caption{Example scene graphs generated by our graph convolution fully-trained model.}
    \label{fig:gen_sg}
\end{figure*}

Figure~\ref{fig:gen_sg} shows more examples of scene graphs generated by our model. The scene graph for the image in the middle of a woman riding a motorcycle shows that our model is able to identify the main action taking place in the image. It is also able to correctly identify parts of the motorcycle, such as \object{seat}, \object{tire}, and \object{light}. The scene graph and image in the bottom right shows that our model can identify parts of the woman's body, like \object{nose} and \object{leg}. It is also able to predict the woman's actions: \predicate{carrying} the \object{bag} and \predicate{holding} the \object{umbrella}.

\subsection{Semantic transformations}
Table~\ref{tab:gcn_sem_viz} shows more examples of semantic transformations applied to the embedding feature space of objects. \object{child} transformed by \predicate{walking on} resembles objects that we walk on: \object{street}, \object{sidewalk}, and \object{snow}. We also learn more specific and rare relationships such as \predicate{attached to}. We observe that \object{sign} transformed by \predicate{attached to} most closely resembles objects such as \object{pole} and \object{fence}.

\begin{table}[t!]
\centering
\caption{We visualize a predicate's semantic transformations by showing the closest objects to a given transformed subject.}
\begin{tabular}{ l l l }
\toprule
  \textbf{subject} & \textbf{object} & \textbf{closest objects} \\
  \midrule
  girl & riding & wave, skateboard, bike, horse \\
  man & wears &  shirt, jacket, hat, cap, helmet \\
  person & has & hair, head, face, arm, ear \\
  dog & laying on & bed, beach, bench, desk, table \\ 
  child & walking on & street, sidewalk, snow, beach \\
  boy & sitting on & bench, bed, desk, chair, toilet \\
  umbrella & covering & kid, people, skier, person, guy \\
  tail & belonging to & cat, elephant, giraffe, dog \\
  stand & over & street, sidewalk, beach, hill \\
  mountain & and & hill, mountain, skier, snow \\
  motorcycle & parked on & street, sidewalk, snow, beach \\
  sign & attached to & pole, fence, shelf, post, building \\
  sidewalk & in front of & building, room, house, fence \\
  kid & watching & giraffe, zebra, plane, horse \\
  men & looking at & airplane, plane, bus, laptop \\
  child & standing on & sidewalk, beach, snow, track \\
  guy & holding & racket, umbrella, glass, bag \\
  motorcycle & has & heel wing handle tire engine \\
  \bottomrule
\end{tabular}
\label{tab:gcn_sem_viz}
\end{table}

\subsection{Visualize object representations}
\begin{figure*}[t!]
    \centering
    \includegraphics[width=0.9\linewidth]{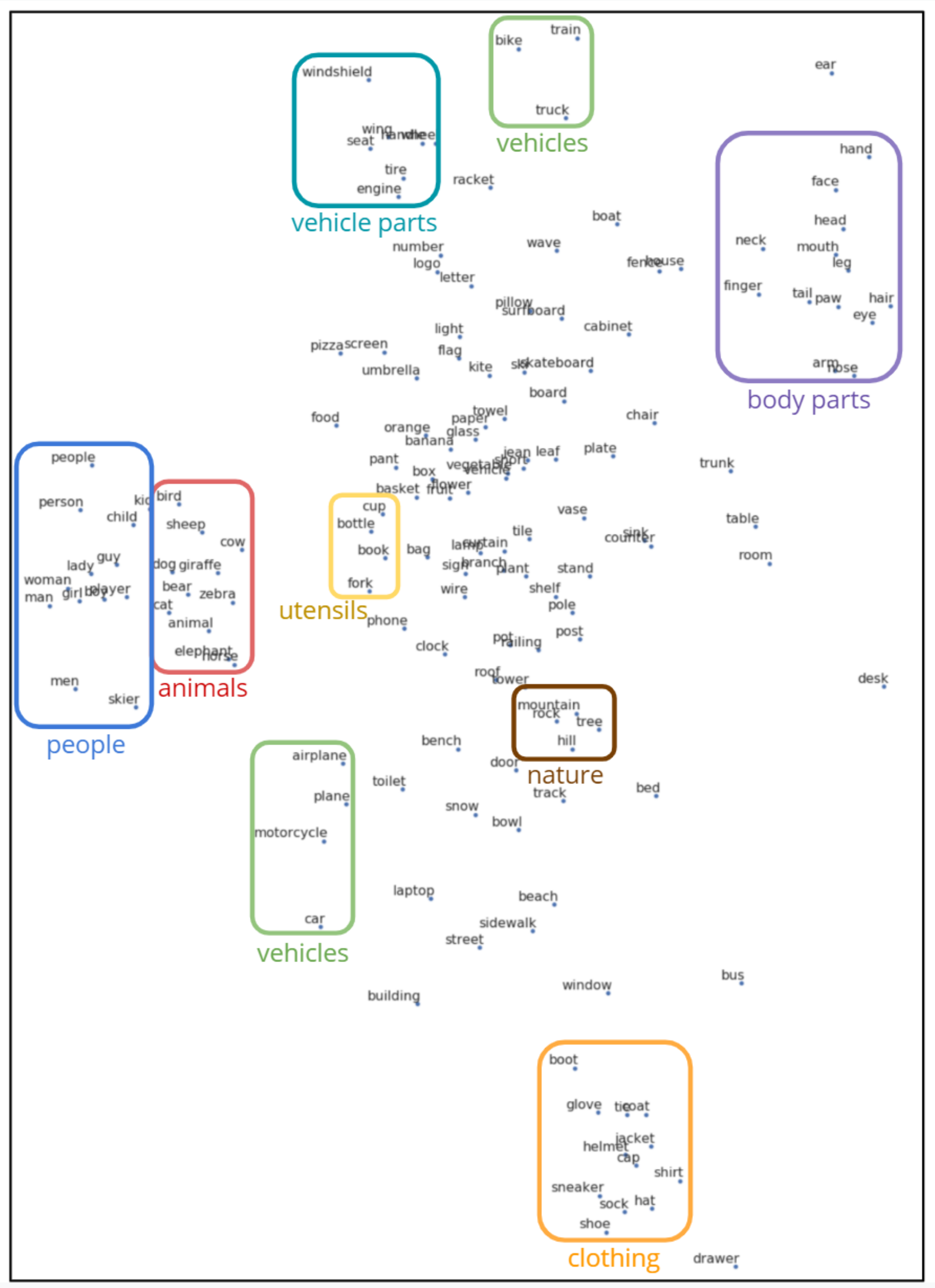}
    \caption{We show a $2$-dimensionl tSNE visualization of the object category embeddings learned by our model.}
    \label{fig:gcn_hidden_repr}
\end{figure*}

In the process of training our predicate functions, we learn representations for each object instance we encounter. From the embedding of each object instance, we calculate the average object category embedding. Each of the $150$ distinct object categories is embedded into a learned $1174$-dimension space. Figure~\ref{fig:gcn_hidden_repr} shows a t-SNE visualization of these embeddings. We observe object categories that participate in similar relationships grouped together. For example, embeddings for \texttt{bird}, \texttt{cow}, \texttt{bear}, and other animals are close together (inside the red rectangle).

\subsection{Inverse predicate functions}
To understand the effect of including inverse predicate functions, we performed an ablation study where the inverse predicate functions were omitted from the model. We found that the semantic-only model trained without inverse functions performed $2.53\%$ worse on recall@$50$ than the semantic model with inverse functions.

\begin{figure*}[t!]
    \centering
    \includegraphics[width=\columnwidth]{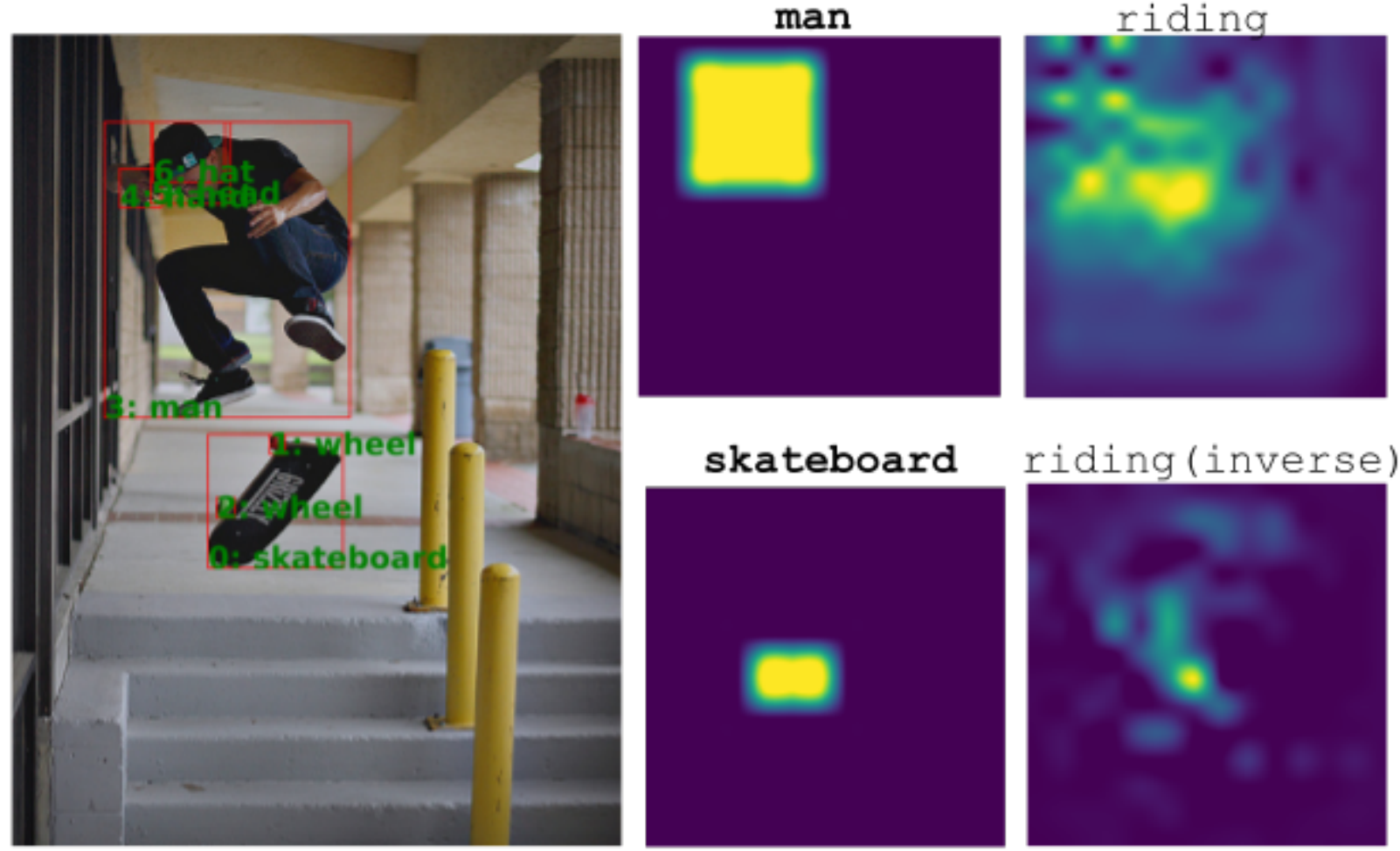}
    \caption{We visualize inverse predicate function transformations.}
    \label{fig:spa_shift_inv_viz}
\end{figure*}

We also visualize how these inverse functions transform a particular subject when compared to the output of the forward function as shown in Figure~\ref{fig:spa_shift_inv_viz}. We observe that the spatial function for the predicate \predicate{riding} shifts attention below the \object{person} in the image. Qualitatively, this is the expected result because the \object{skateboard} is below the \object{person}. The inverse transformation of \predicate{riding} shifts the \object{skateboard} mask slightly above the skateboard. Similarly, this is also the expected result because skateboarders are typically above their boards.

\begin{table}[t!]
\centering
\caption{Comparing performance varying the number of GCN layers. We find that $2$ layers performs best but the change in performance is minimal when we change the number of layers.}
\begin{tabular}{l | l}
\toprule
  \textbf{\# GCN layers} & \textbf{Recall@50}\\
  \midrule
  1 & 74.26\\
  2 & 75.15\\
  3 & 74.26\\
  4 & 74.78\\
  \bottomrule
\end{tabular}
\label{tab:gcn_layers}
\end{table}

\begin{table}[t!]
\centering
\caption{Comparing performance varying the depth of the semantic model. We find that $4$ performs best and increasing the depth of the model leads to overfitting.}
\begin{tabular}{l | l}
\toprule
  \textbf{\# Semantic model depth} & \textbf{Recall@50}\\
  \midrule
  2 & 73.75\\
  4 & 75.33\\
  5 & 73.75\\
  \bottomrule
\end{tabular}
\label{tab:semantic_layers}
\end{table}

\begin{table}[t!]
\centering
\caption{Comparing performance varying hidden size of the semantic model. Performance of the model increases as we increase hidden size but the model needs to be regularized more to prevent overfitting.}
\begin{tabular}{l | l}
\toprule
  \textbf{Semantic model hidden size} & \textbf{Recall@50}\\
  \midrule
  200 & 74.35\\
  300 & 73.76\\
  400 & 75.33\\
  \bottomrule
\end{tabular}
\label{tab:semantic_hidden}
\end{table}

\begin{table}[t!]
\centering
\caption{Comparing performance varying the size of the semantic object representations. We find that $1024$ is the best choice.}
\begin{tabular}{l | l}
\toprule
  \textbf{Object representation size} & \textbf{Recall@50}\\
  \midrule
  256 & 74.54\\
  512 & 74.94\\
  1024 & 75.64\\
  2048 & 75.33\\
  \bottomrule
\end{tabular}
\label{tab:obj_size}
\end{table}

\begin{table}[t!]
\centering
\caption{Comparing performance varying the backbone architecture used to extract representations for objects. We find that ResNet only performs marginally better than VGG.}
\begin{tabular}{l | l}
\toprule
  \textbf{Backbone} & \textbf{Recall@100}\\
  \midrule
  ResNet50 & 0.2289\\
  VGG & 0.2094\\
  \bottomrule
\end{tabular}
\label{tab:backbone}
\end{table}

\subsection{Ablations}
We show how our GCN model performance differs based on various model ablations. First, in Table~\ref{tab:gcn_layers}, we test how our model performs when the number of GCN layers is changed from $1$ to $4$. We find that $2$ GCN layers gives us the best performance, although using $1$, $3$, or $4$ GCN layers does not greatly affect recall@$50$. 

Next, in Table~\ref{tab:semantic_layers}, we vary the depth, i.e.~the number of layers, in each semantic function. We find that having $4$ layers achieves the best performance, while having more than $4$ layers leads to overfitting. 

Next, in Table~\ref{tab:semantic_hidden}, we vary the hidden size of the semantic model and find that a hidden size of $400$ performs best but needs more regularization. Similarly, in Table~\ref{tab:obj_size}, We also experimented with different object representation sizes and found that having a representation of size $1024$ worked the best. Other representation sizes are between $0.3$ and $1.1$ recall@50 lower. 

Lastly, in Table~\ref{tab:backbone}, we compare how different object detection backbones affect few-shot performance. We find that using a ResNet50 backbone achieves better performance than using the VGG object detection architecture. However, the increase in performance is minimal. Even with VGG backbone, our few-shot performance is higher than all existing baselines.

\end{document}